# Evaluating the Performance of Open-Vocabulary Object Detection in Low-quality Image

**Po-Chih Wu**
National Yang Ming Chiao Tung University
Hsinchu City, Taiwan
gohakushi@gmail.com

## Abstract

Open-vocabulary object detection enables models to localize and recognize objects beyond a predefined set of categories and is expected to achieve recognition capabilities comparable to human performance. In this study, we aim to evaluate the performance of existing models on open-vocabulary object detection tasks under low-quality image conditions. For this purpose, we introduce a new dataset that simulates low-quality images in the real world. In our evaluation experiment, we find that although open-vocabulary object detection models exhibited no significant decrease in mean average precision (mAP) scores under low-level image degradation, the performance of all models dropped sharply under high-level image degradation. OWLv2 models consistently performed better across different types of degradation, while OWL-ViT, Grounding-DINO, and Detic showed significant performance declines. The image dataset used in this study, along with image processing code, is available at https://github.com/gohakushi1118/Low-quality-image-dataset

## 1. Introduction

Open-vocabulary object detection models use vision language pretraining to detect unseen objects from free-text queries, making them useful for applications such as robotics [1] and autonomous driving [2]. Most open-vocabulary object detection models are evaluated on high-quality datasets such as MS COCO [3] and LVIS [4]. However, images captured in the real-world environment often suffer from degradation such as motion blur, sensor noise, and overexposure. Therefore, to explore the impact of open-vocabulary object detection models under low-quality conditions, we propose a new benchmark [5] based on the COCO dataset, in which images are processed into four categories, including lossy compression, image intensity, image noise, and image blur. We evaluate six state-of-the-art open-vocabulary object detection models, and our results reveal that severe degradation leads to significant performance drops across all models. Models also show different levels of sensitivity to specific low-quality types. In summary, our contribution to this work is as follows:

1. We construct a new dataset, the Low-quality image dataset, containing four degradation types.
2. We evaluate six open-vocabulary object detection models on this benchmark. Experiments show that their sensitivity varies depending on the type and level of degradation.

## 2. Related work

### 2.1 Open-vocabulary object detection

Traditional object detection methods struggle to classify and identify unseen objects at inference time. In recent years, open-vocabulary object detection has emerged as a new tend in advanced detectors. Open-vocabulary object detection models can detect objects without requiring bounding box annotations for the target classes during training, and it achieves detection based on free-text queries. ViLD [6] first introduced the use of CLIP [7] embeddings to align image regions with textual class names. RegionCLIP [8] enhanced this by aligning region-level features with text, improving detection precision. GLIP [9] employed large-scale image-text pretraining to support zero-shot detection. OWL-ViT [10] extended this approach with a Vision Transformer architecture capable of open-world object localization via text queries.

### 2.2 Microsoft COCO

The MS COCO (Microsoft Common Objects in Context) dataset is widely used for object detection, segmentation, and captioning tasks. It consists of two main versions: MS COCO 2014, which contains 83K training images, 41K validation images, and 41K test images with bounding boxes labeled for 80 categories and MS COCO 2017, which contains 118K training images, 5K validation images, and 41K test images with bounding boxes labeled for 80 categories. This dataset has become a benchmark standard for evaluating the performance of models.

# 3. Methodology

### 3.1 Dataset

Due to the scarcity of suitable image datasets online related to low-quality images, we created a new dataset for this study. Low-quality image dataset is based on the COCO 2017 validation set, with images processed into four categories, including lossy compression, image intensity, image noise, and image blur. In total, the dataset comprises 100,000 processed images. Each low-quality category includes five levels of image degradation with 5,000 processed images per level.

### 3.2 Model

We evaluate the performance of six different open-vocabulary object detection models. Two of the models, developed by Google, are based on the OWL-ViT framework: OWL-B/16 and OWL-B/32. Another two models, also from Google, are based on the OWLv2 [11] framework: OWLv2-B/16 and OWLv2-L/14. The remaining two models are GroundingDINO [12] and Detic [13]. These are advanced open-source models currently available on Hugging Face.

### 3.3 Image processing

**Gaussian noise** is a type of random noise that follows a Gaussian distribution, and it simulates the random variations in pixel values.

$$I' = I + G(0, \sigma^2)$$

where $I'$ is the image with Gaussian noise, $I$ is the original image, and $G(0, \sigma^2)$ represents a Gaussian distribution with a mean of zero and a standard deviation of $\sigma$. In this study, we set the standard deviation value to $[10, 20, 30, 40, 50]$.

**Average blur** is a low-pass filter that smooths an image by replacing each pixel with the average value of its neighboring pixels within a defined kernel value, and it roughly simulates several blur conditions such as motion blur or focus blur.

$$I'(x, y) = \frac{1}{k^2} \sum_{(r,c) \in S_{xy}} I(r, c)$$

where $I'(x, y)$ is the image with average blur, $I(r, c)$ is the original image, $k$ is the kernel value in both row and column, and $S_{xy}$ represents the set of coordinates in $k \times k$ neighborhood centered at pixel $(x, y)$. In this study, we set the kernel value to $[4, 6, 8, 10, 12]$.

**Gamma correction** is a method for adjusting image intensity through nonlinear transformation, used to simulate underexposure or overexposure scenarios.

$$I' = 255 \cdot \left(\frac{I}{255}\right)^{\frac{1}{\gamma}}$$

where $I'$ is the image with gamma correction, $I$ is the original image, and $\gamma$ is the gamma value. The gamma value of $\gamma < 1$ darkens the image, while the value of $\gamma > 1$ brightens it. The pixel value is normalized to the $[0, 1]$ by dividing by 255, then transformed using gamma correction, and rescaled back to $[0, 255]$. In this study, we set the gamma value to $[0.4, 0.8, 1.2, 1.6, 2.0]$.

**Lossy compression** method used in this study is JPEG [14] compression, which operates by transforming image data into the frequency domain using the Discrete Cosine Transform. All images are converted to JPEG format, and the compression strength is controlled by quality values. In our study, we set the quality values to [0, 20, 40, 60, 80].

# 4. Experiments

### 4.1 Experimental setup

We construct the evaluation label set from 80 standard COCO categories, represented as text prompts. During inference, each image is combined with all category names processed by a multi-modal encoder that extracts both text and image features. The model outputs bounding boxes and confidence scores. A confidence threshold of 0.1 is applied to filter out low-confidence predictions. The predicted label is matched to its corresponding COCO category ID. To convert the predictions into the COCO annotation format, each predicted bounding box is transformed from corner coordinates $(x_{min}, y_{min}, x_{max}, y_{max})$ to COCO format $(x, y, width, height)$. We use the mAP metric with IoU thresholds from 0.50 to 0.95 in steps of 0.05 for performance evaluation.

### 4.2 Results

OWLv2 models consistently demonstrated the most stable performance across all conditions, with OWLv2-L/14 achieving the best results in most cases. GroundingDINO showed moderate tolerance to degradation. In contrast, OWL-ViT models and Detic experienced significant performance declines under severe conditions, indicating their higher sensitivity to image quality degradation.

| Detector | Quality | | | | | |
|---|---|---|---|---|---|---|
| | 0 | 20 | 40 | 60 | 80 | Original |
| OWL$_{(B/32)}$ | 1.0 | 22.2 | 20.7 | 19.4 | 20.1 | 22.0 |
| OWL$_{(B/16)}$ | 0.6 | 15.5 | 16.9 | 18.1 | 20.1 | 24.9 |
| OWLv2$_{(B/16)}$ | 4.2 | 32.4 | 34.9 | 35.7 | **36.5** | 31.2 |
| OWLv2$_{(L/14)}$ | **8.6** | **34.8** | **36.2** | **36.4** | 36.3 | **34.3** |
| GDino$_{(Tiny)}$ | 1.6 | 23.5 | 27.6 | 28.9 | 30.4 | 31.3 |
| Detic | 0.2 | 16.7 | 20.3 | 21.4 | 22.0 | 22.8 |

Table 1. Results of performance under different levels of lossy compression. The highest performance among all models is highlighted in **bold**.

| Detector | Gamma Value | | | | | |
|---|---|---|---|---|---|---|
| | 0.4 | 0.8 | 1.2 | 1.6 | 2.0 | Original |
| OWL $_{(B/32)}$ | 17.2 | 19.8 | 20.1 | 19.8 | 19.1 | 22.0 |
| OWL $_{(B/16)}$ | 18.6 | 22.0 | 22.1 | 22.0 | 21.2 | 24.9 |
| OWLv2$_{(B/16)}$ | 33.7 | **36.7** | **36.9** | **36.6** | **36.4** | 31.2 |
| OWLv2$_{(L/14)}$ | **33.8** | 35.2 | 35.3 | 35.3 | 35.3 | **34.3** |
| GDino $_{(Tiny)}$ | 28.5 | 31.4 | 31.2 | 30.8 | 30.4 | 31.3 |
| Detic | 20.4 | 22.5 | 22.4 | 22.3 | 22.1 | 22.8 |

Table 2. Results of performance under different levels of image intensity. The highest performance among all models is highlighted in **bold**.

| Detector | Standard Deviation (SD) | | | | | |
|---|---|---|---|---|---|---|
| | 10 | 20 | 30 | 40 | 50 | Original |
| OWL$_{(B/32)}$ | 23.1 | 20.0 | 17.0 | 14.3 | 11.9 | 22.0 |
| OWL$_{(B/16)}$ | 23.3 | 19.2 | 15.2 | 11.8 | 9.0 | 24.9 |
| OWLv2$_{(B/16)}$ | 35.3 | 31.7 | 27.7 | 23.7 | 19.9 | 31.2 |
| OWLv2$_{(L/14)}$ | **35.7** | **34.7** | **32.9** | **30.6** | **28.2** | **34.3** |
| GDino$_{(Tiny)}$ | 29.9 | 27.3 | 24.2 | 21.3 | 18.1 | 31.3 |
| Detic | 21.4 | 18.6 | 15.1 | 11.6 | 8.5 | 22.8 |

Table 3. Results of performance under different levels of image noise. The highest performance among all models is highlighted in **bold**.

| Detector | Kernel value | | | | | |
|---|---|---|---|---|---|---|
| | 4 | 6 | 8 | 10 | 12 | Original |
| OWL$_{(B/32)}$ | 18.8 | 16.7 | 14.3 | 12.2 | 10.2 | 22.0 |
| OWL$_{(B/16)}$ | 19.0 | 13.1 | 15.8 | 11.4 | 9.4 | 24.9 |
| OWLv2$_{(B/16)}$ | 33.9 | 30.7 | 27.8 | 25.0 | 22.5 | 31.2 |
| OWLv2$_{(L/14)}$ | **35.2** | **33.5** | **31.0** | **28.4** | **26.0** | **34.3** |
| GDino$_{(Tiny)}$ | 28.7 | 25.2 | 21.4 | 18.1 | 15.1 | 31.3 |
| Detic | 19.8 | 17.2 | 14.2 | 11.1 | 8.7 | 22.8 |

Table 4. Results of performance under different levels of image blur. The highest performance among all models is highlighted in **bold**.

## 5. Conclusion

This paper evaluates six open-vocabulary object detection models under various low-quality conditions. To support this study, we introduce a new dataset based on the COCO 2017 validation set, containing 100,000 low-quality images. Our experiments reveal that all models suffer significant performance drops under severe degradation. The results also reveal that different models have different levels of sensitivity to each type of low-quality condition. Future work should explore training strategies and architectures that improve models under visual challenges in the real world.